%% file: main.tex
\documentclass[10pt,twocolumn,letterpaper]{article}

\usepackage{3dv}
\usepackage{times}
\usepackage{epsfig}
\usepackage{graphicx}
\usepackage{amsmath}
\usepackage{amssymb}

\usepackage{array}
\usepackage{algorithm}
\usepackage{algpseudocode}
\usepackage{multirow}
\usepackage{pifont}
\usepackage{subcaption}
\input{custom/acros}
\input{custom/commands}

\usepackage[pagebackref=true,breaklinks=true,colorlinks,bookmarks=false]{hyperref}

\threedvfinalcopy 

\def\threedvPaperID{121} 

\ifthreedvfinal\fi
\begin{document}

\title{ManiFlow: Implicitly Representing Manifolds with Normalizing Flows}

\author{Janis Postels$^1$ \hspace{8mm} Martin Danelljan$^1$ \hspace{8mm} Luc Van Gool$^1$ \hspace{8mm} Federico Tombari$^{2,3}$\\
$^1$ETH Zurich \hspace{8mm} $^2$Google \hspace{8mm} $^3$Technical University Munich\\
{\tt\small \{jpostels, martin.danelljan, vangool\}@vision.ee.ethz.ch \hspace{4mm} tombari@in.tum.de}
}

\maketitle

\input{sections/abstract}

\section{Introduction}

\input{sections/introduction}

\section{Related Work}

\input{sections/related_work}

\section{Method}

\input{sections/method}

\section{Experiments}

\input{sections/experiments}

\section{Conclusion}

\input{sections/conclusion}



\clearpage
%
%

{\small
\bibliographystyle{ieee_fullname}
\bibliography{egbib}
}

\end{document}

%% file: custom/acros.tex
\usepackage{acro}

\DeclareAcronym{nf}{
  short = NF ,
  long = Normalizing Flow
}

\DeclareAcronym{nn}{
  short = NN ,
  long = Neural Network
}

\DeclareAcronym{dpm}{
  short = DPM ,
  long = Diffusion Probabilistic Model
}

\DeclareAcronym{cd}{
  short = CD ,
  long = Chamfer Distance
}

\DeclareAcronym{emd}{
  short = EMD ,
  long = Earth Mover Distance
}

%% file: custom/commands.tex
\newcommand{\secref}[1]{Sec. \ref{#1}}
\newcommand{\eref}[1]{Eq. \ref{#1}}
\newcommand{\tabref}[1]{Tab. \ref{#1}}
\newcommand{\figref}[1]{Fig. \ref{#1}}
\newcommand{\aref}[1]{Alg. \ref{#1}}

\newcommand{\figsize}{0.2}

%% file: sections/abstract.tex
\begin{abstract}
    
    Normalizing Flows (NFs) are flexible explicit generative models that have been shown to accurately model complex real-world data distributions. However, their invertibility constraint imposes limitations on data distributions that reside on lower dimensional manifolds embedded in higher dimensional space. Practically, this shortcoming is often bypassed by adding noise to the data which impacts the quality of the generated samples. In contrast to prior work, we approach this problem by generating samples from the original data distribution given full knowledge about the perturbed distribution and the noise model. To this end, we establish that NFs trained on perturbed data implicitly represent the manifold in regions of maximum likelihood. Then, we propose an optimization objective that recovers the most likely point on the manifold given a sample from the perturbed distribution. Finally, we focus on 3D point clouds for which we utilize the explicit nature of NFs, i.e. surface normals extracted from the gradient of the log-likelihood and the log-likelihood itself, to apply Poisson surface reconstruction to refine generated point sets.
\end{abstract}

%% file: sections/introduction.tex



The goal of generative modeling is to grasp the fundamental laws governing a data distribution in order to autonomously create alternative realizations. As such, it represents a core element of intelligence and, thus, also machine learning research. Beyond fundamental research, recent advances in generative modeling~\cite{kingma2013auto,goodfellow2014generative,dinh2016density,song2019generative,ho2020denoising} have lead to a variety of real-world applications~\cite{chen2018unsupervised,lugmayr2020srflow,arroyo2021variational}. Existing generative models can be subdivided into implicit and explicit models. Both types aim to generate novel realizations consistent with prior experience. However, unlike their implicit counterparts, explicit models can also attach a likelihood to new observations. This directly gives rise to more applications, such as anomaly detection~\cite{chen2018unsupervised}. Among explicit generative models, \acp{nf}~\cite{dinh2016density} have recently emerged as a family of parametric models of exceptional flexibility. Based on invertible \acp{nn}, they allow modeling high-dimensional data while maintaining the advantage of explicit likelihood evaluation~\cite{kingma2018glow,grathwohl2018ffjord,behrmann2019invertible} and have been successfully applied in a variety of contexts - e.g.\ image super-resolution~\cite{lugmayr2020srflow}, modeling 3D point clouds~\cite{yang2019pointflow} and physical simulation~\cite{kanwar2020equivariant}.

\begin{figure*}[t!]
\begin{center}

    \renewcommand{\figsize}{0.17}
    
    \begin{tabular}{ c c c | c c c }
    
        \centering 
    
        & GT & NF & GT + Noise & NF & NF + LLM \\
        
      \rotatebox[x=0pt,y=25pt]{90}{Uniform}  &
      \includegraphics[width=\figsize\linewidth]{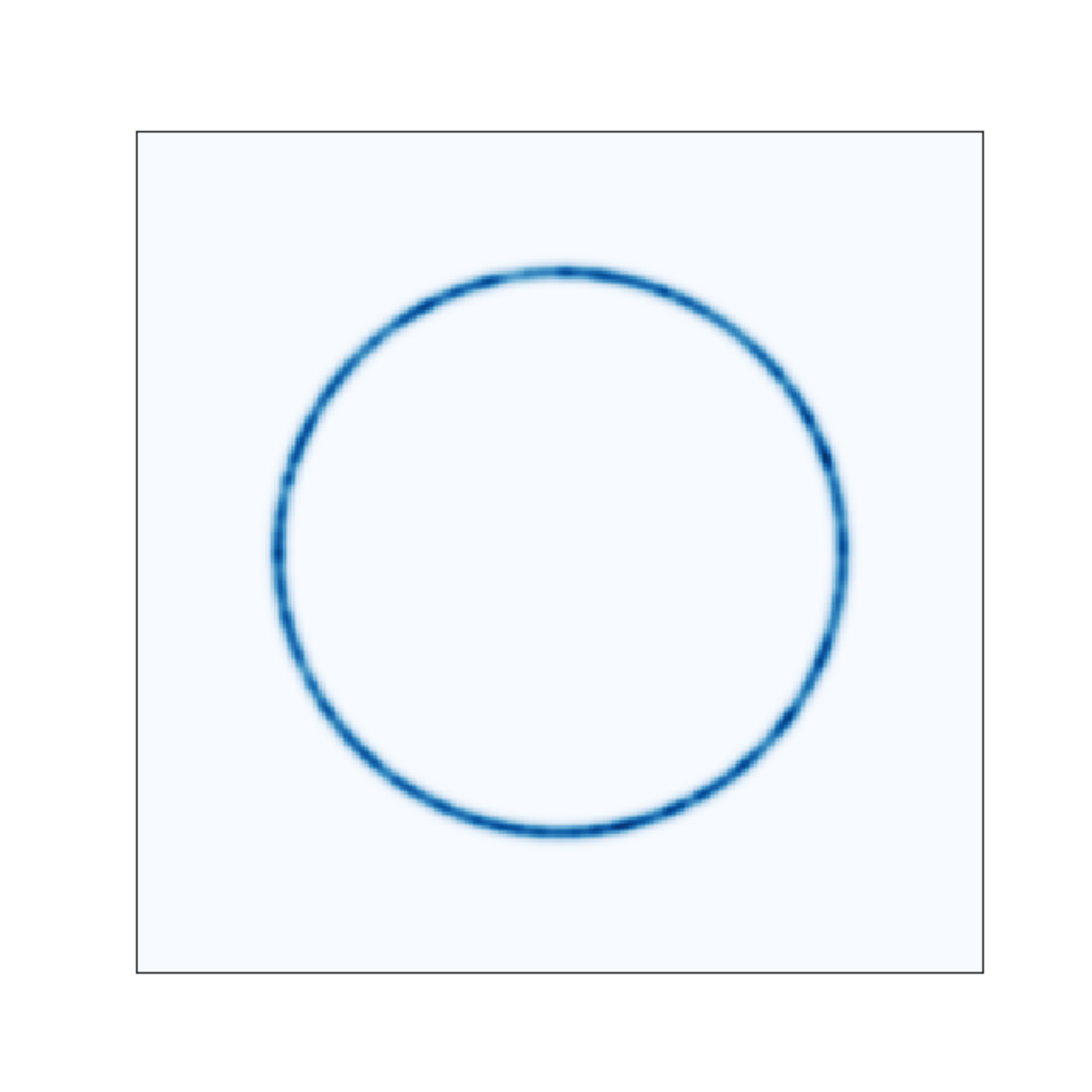} & \includegraphics[width=\figsize\linewidth]{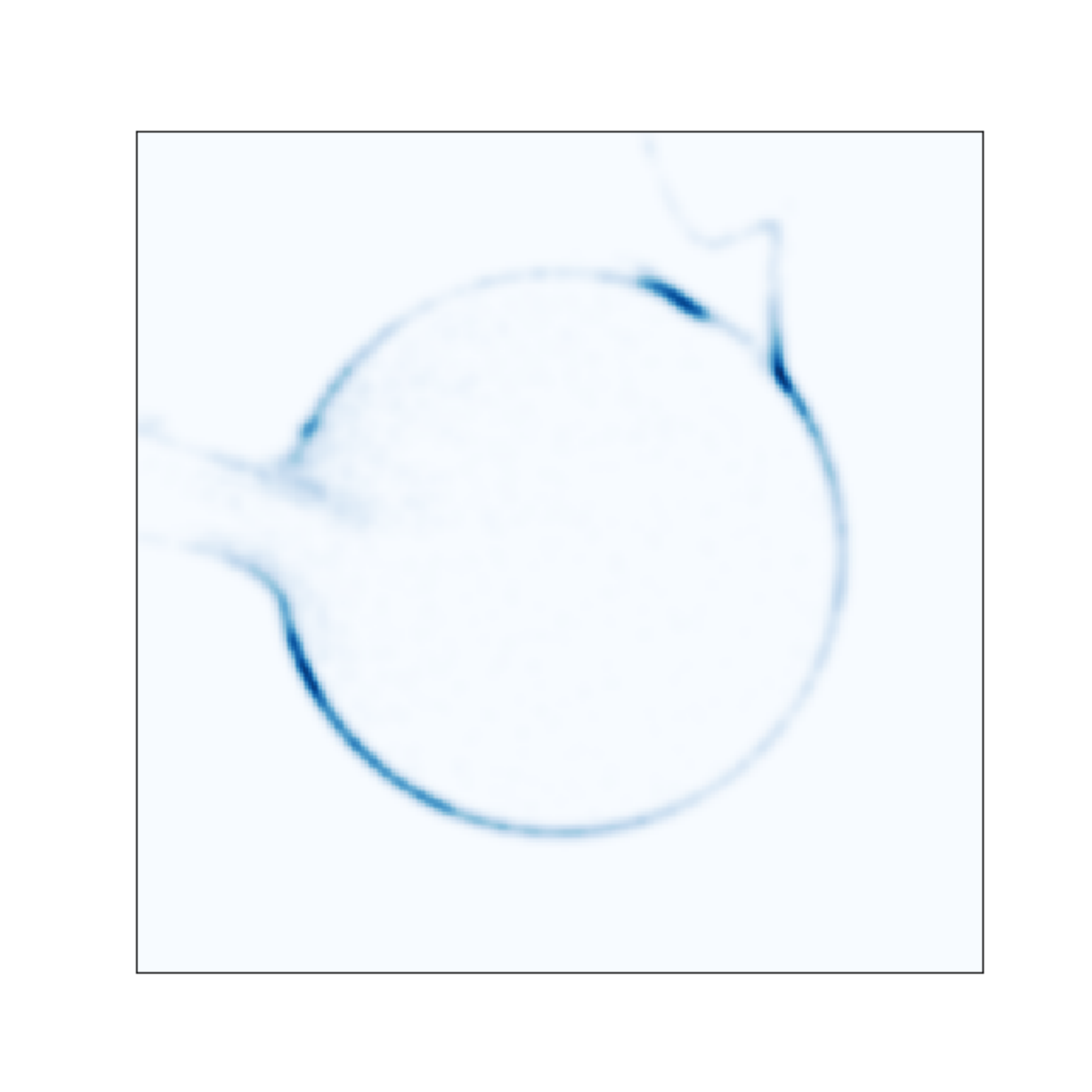} & \includegraphics[width=\figsize\linewidth]{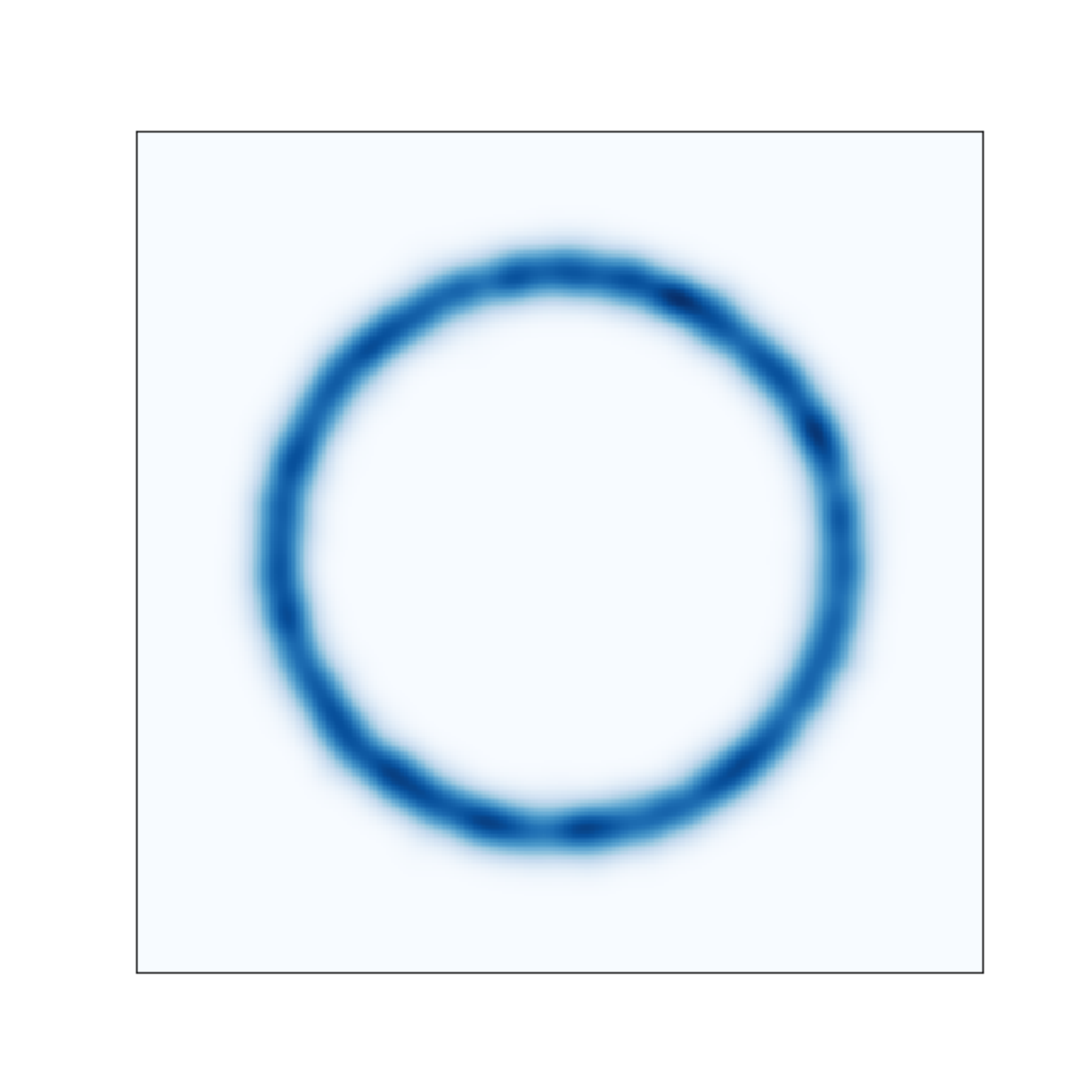} & \includegraphics[width=\figsize\linewidth]{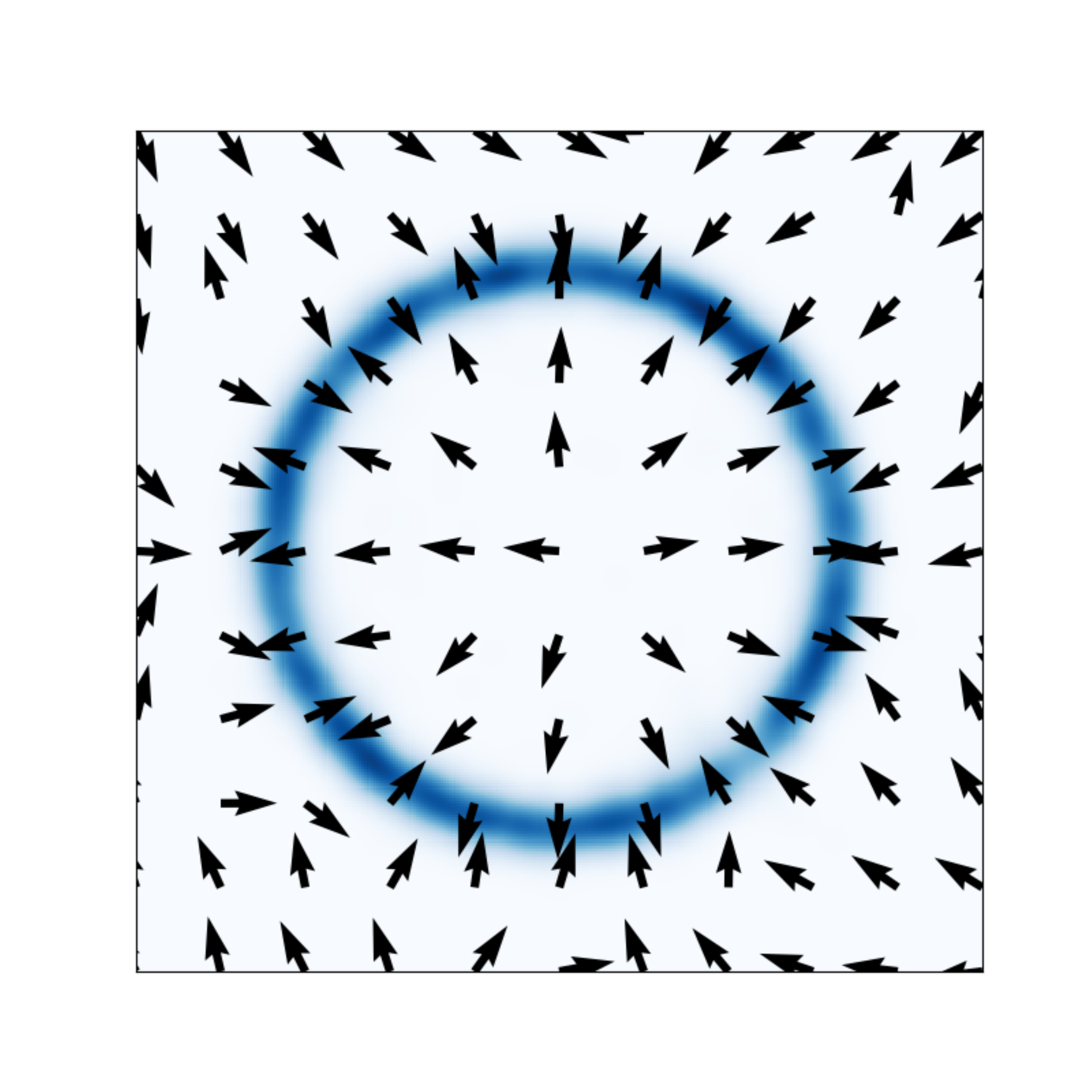} & \includegraphics[width=\figsize\linewidth]{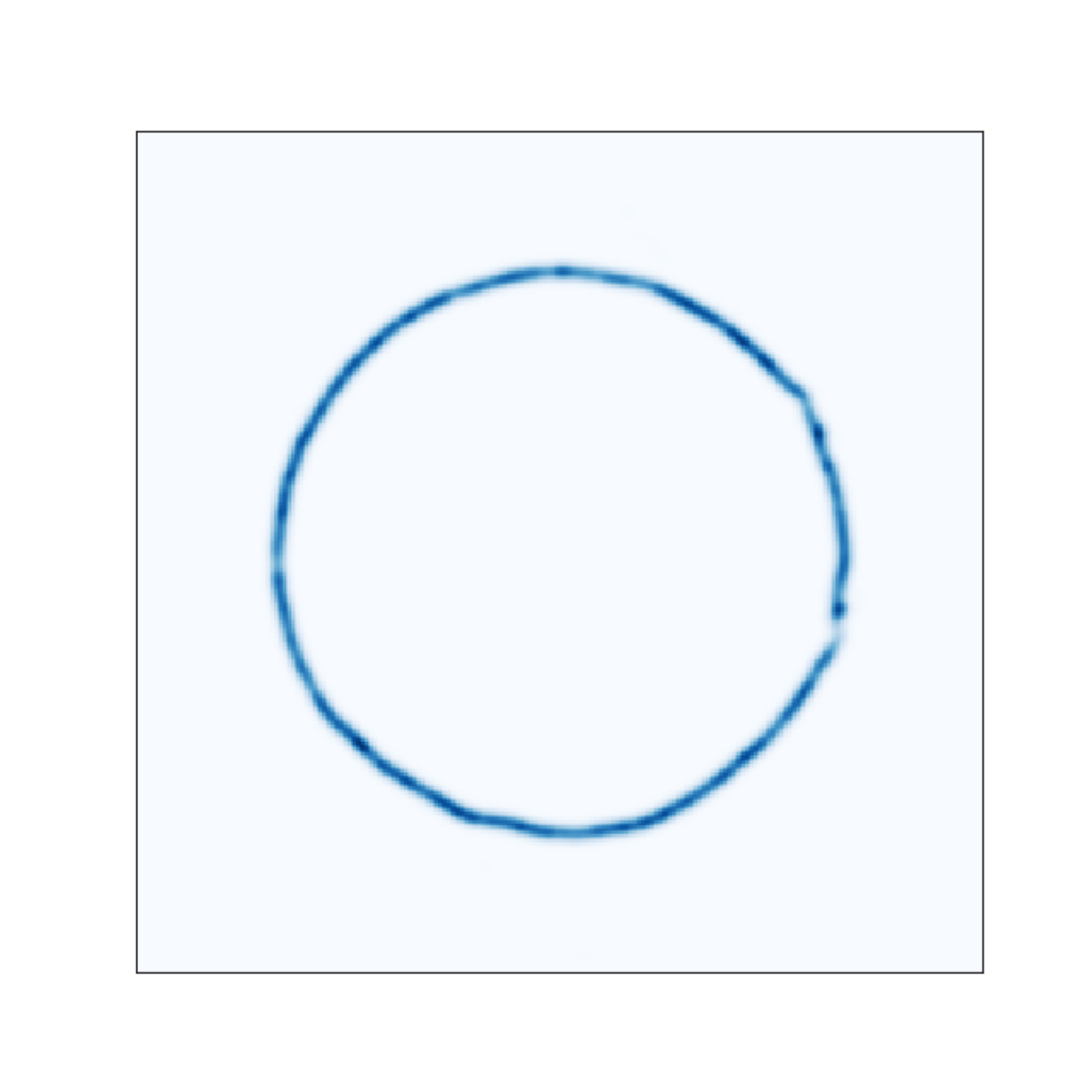} \\
      
      \rotatebox[x=0pt,y=15pt]{90}{Non-Uniform}  &
      \includegraphics[width=\figsize\linewidth]{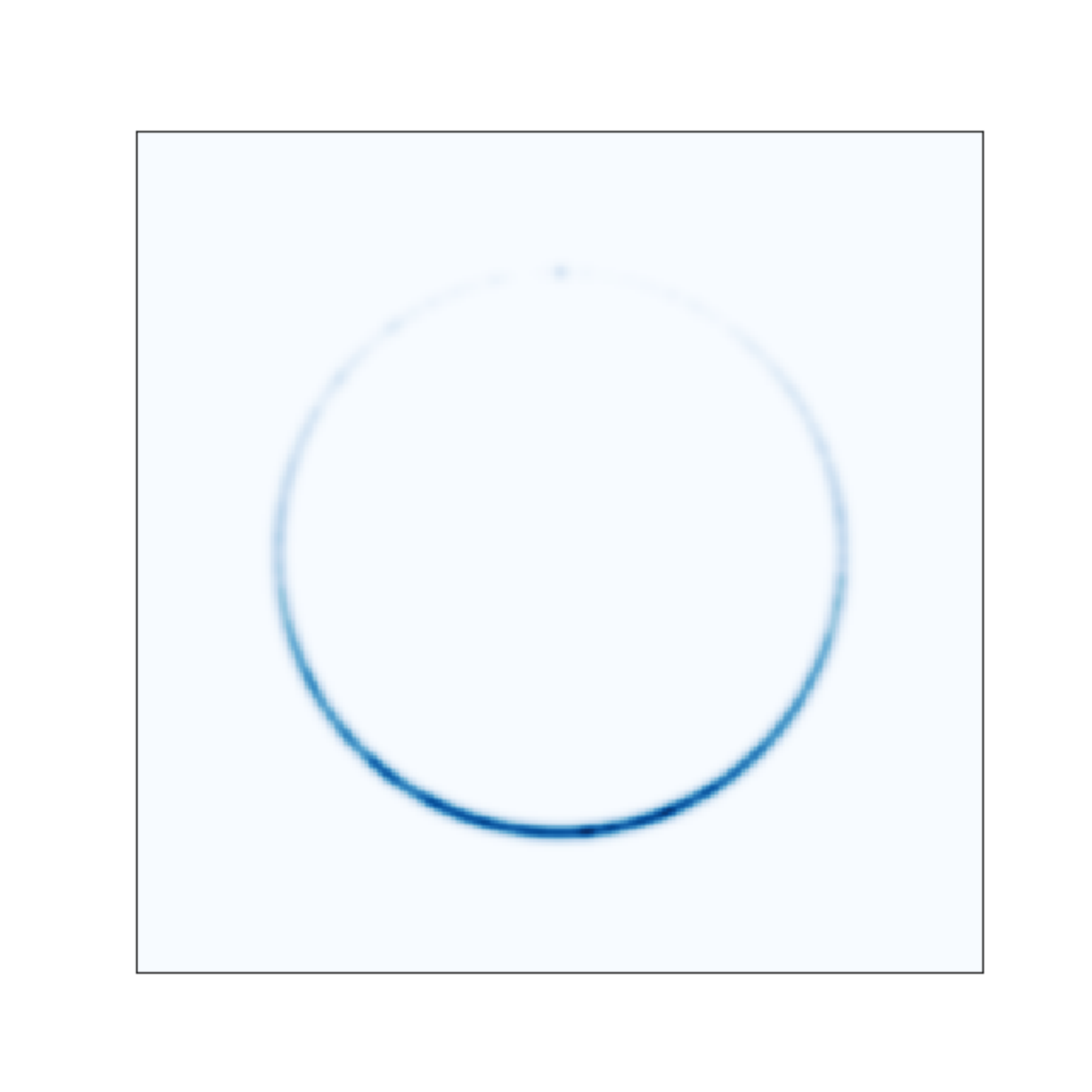} & \includegraphics[width=\figsize\linewidth]{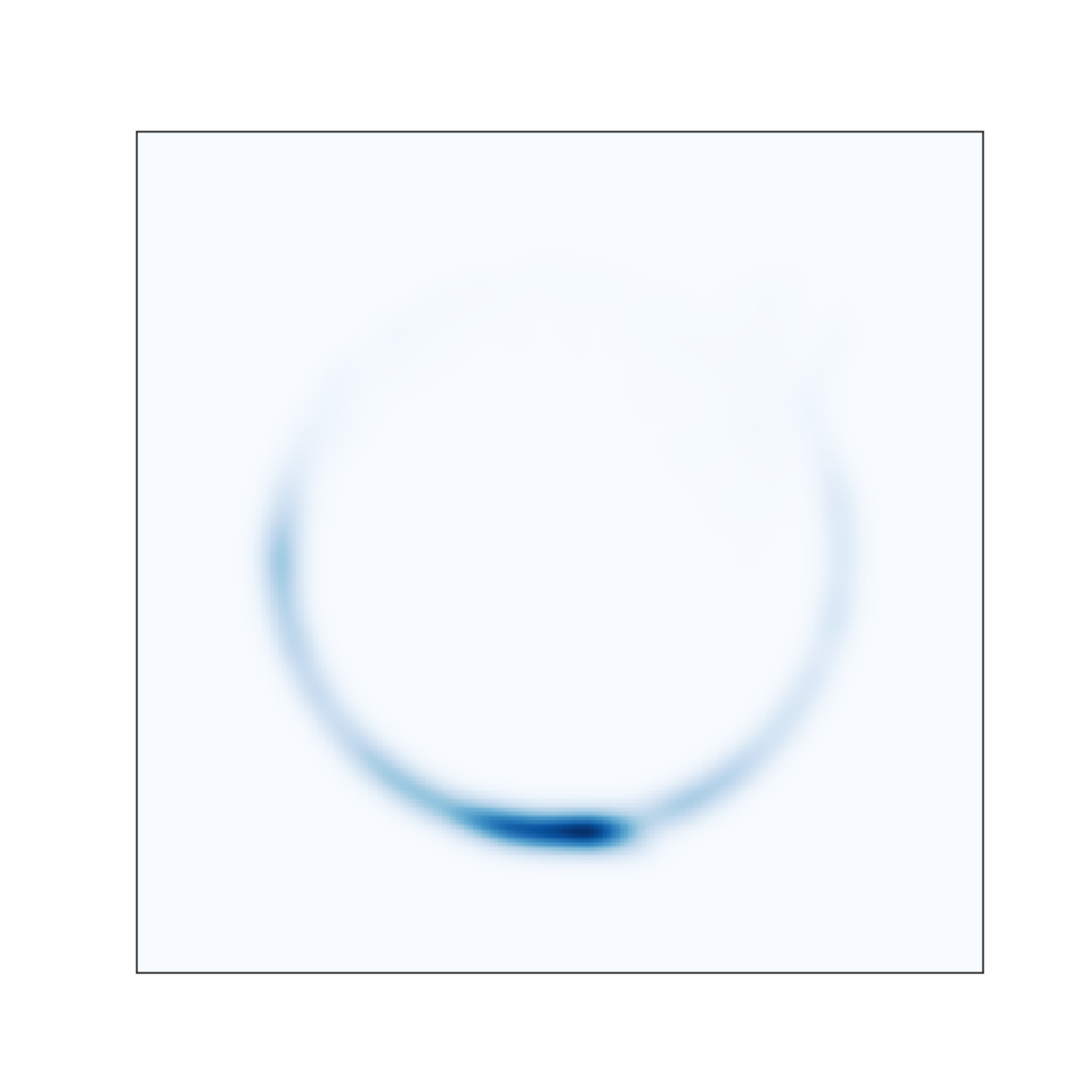} & \includegraphics[width=\figsize\linewidth]{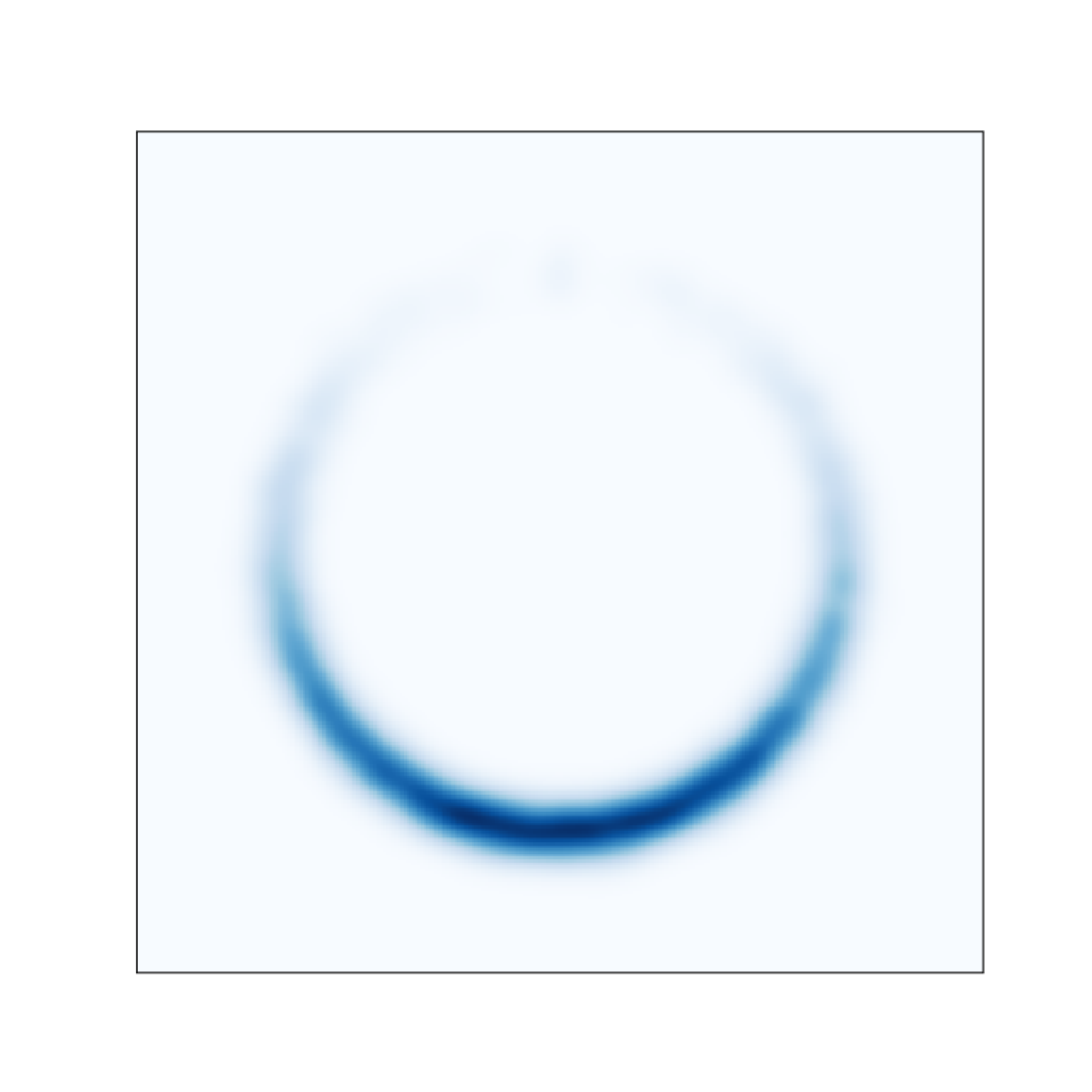} & \includegraphics[width=\figsize\linewidth]{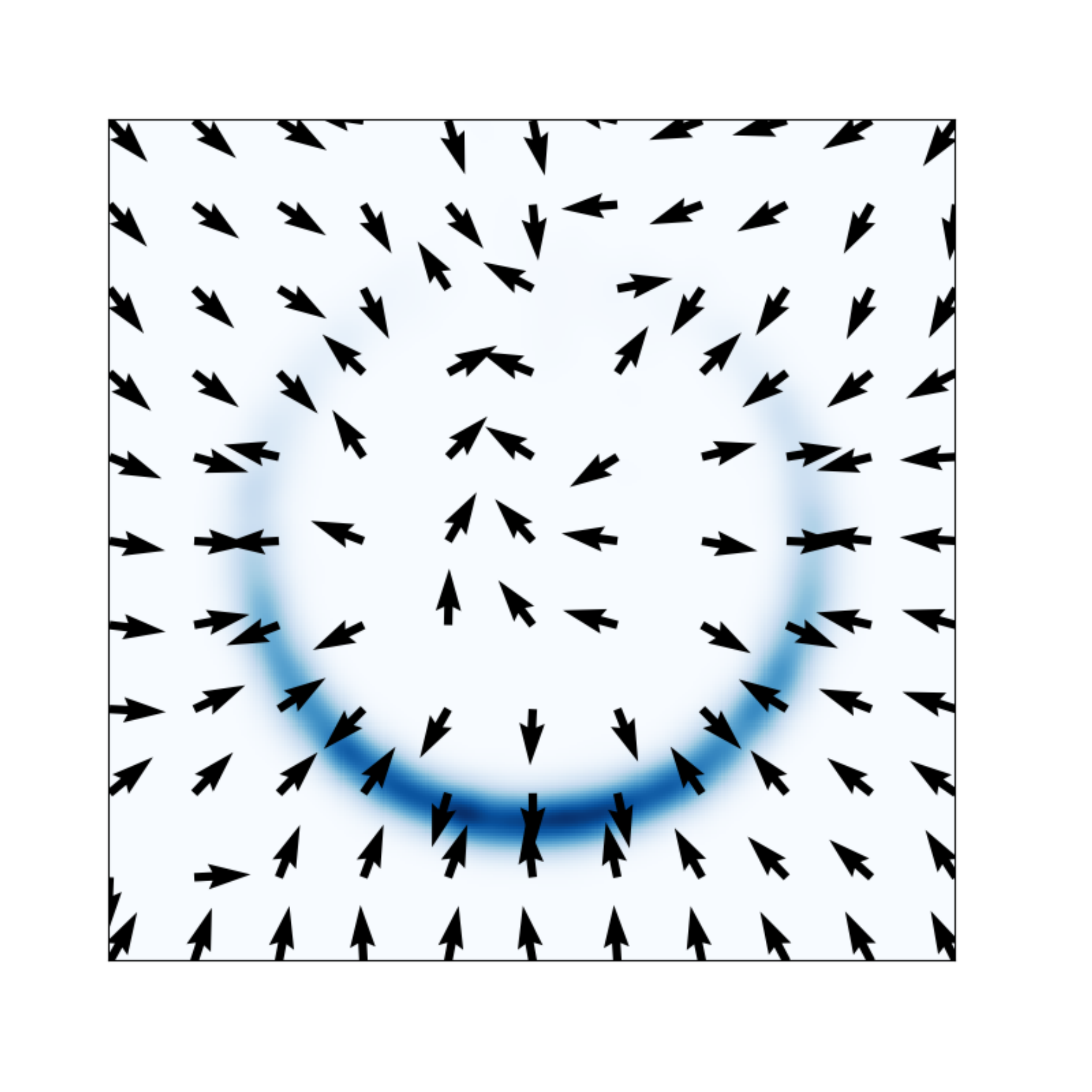} & \includegraphics[width=\figsize\linewidth]{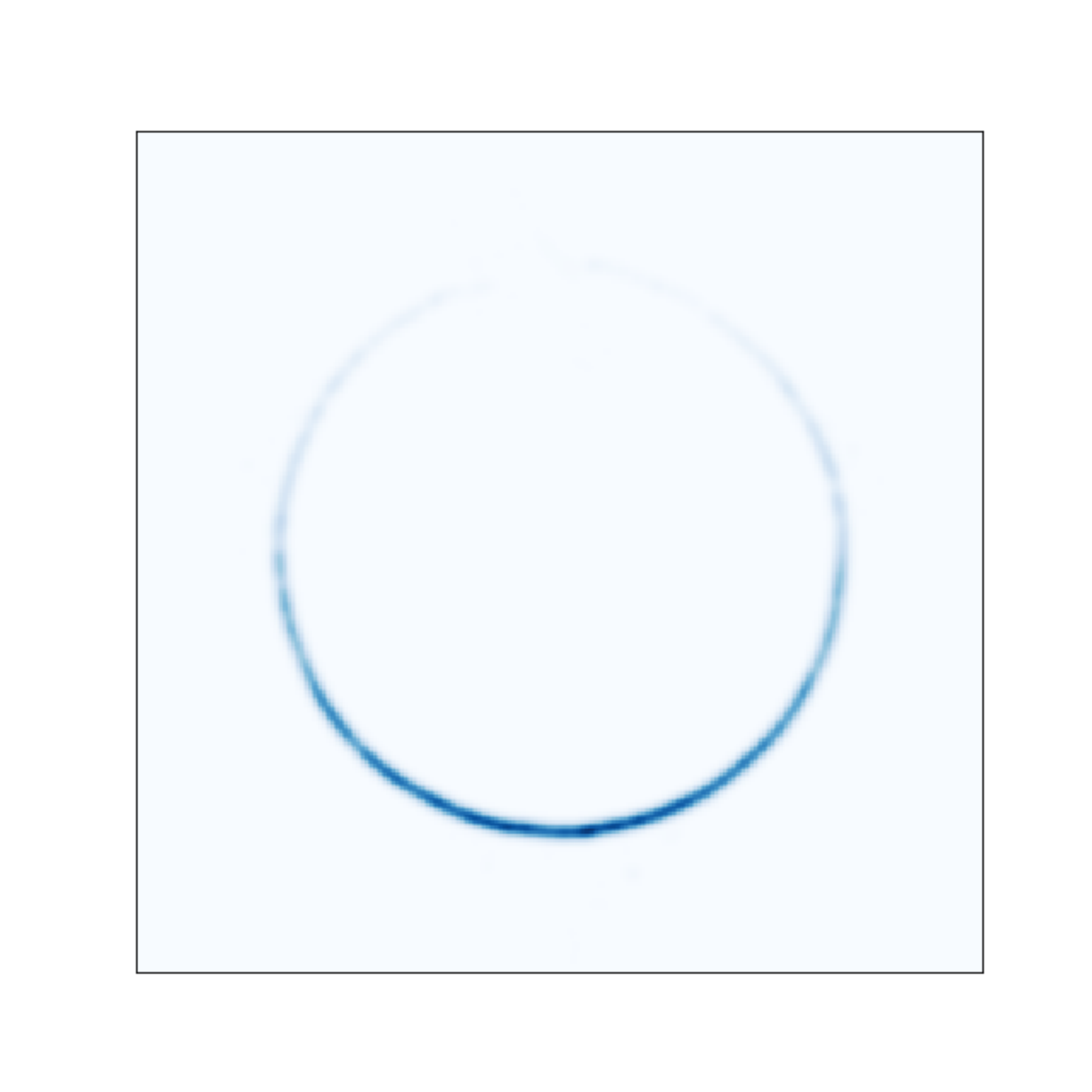} \\
      
    \end{tabular}
    
\end{center}
\caption{In this toy example, we learn a distribution, both uniform (GT) and non-uniform (GT + Noise), that exists on a sphere using a \ac{nf}. Training a vanilla RealNVP~\cite{dinh2016density} on uncorrupted data fails (left). The \ac{nf} is able to learn the noise-perturbed distribution. Applying our proposed log-likelihood maximization (LLM) according to \eref{eq:optimization_criterion}, we are able to recover the original distribution - in the uniform as well as non-uniform scenario. We also visualize the direction of the gradients of the log-likelihood. We observe that they are perpendicular to the manifold in both cases.}\label{fig:toy_example}
\end{figure*}

Despite recent successes, \acp{nf} still possess a fundamental drawback rooted in their core idea - their invertibility. Real-world data often resides on a lower dimensional manifold embedded in higher dimensional space. While apparent for surfaces of 3D shapes, this also holds for other data modalities, e.g.\ images. \acp{nf} rely on invertible \acp{nn} to quantify the likelihood of data instances and are trained by directly maximizing the likelihood of the data. Since a lower dimensional manifold has no volume, \acp{nf} can predict arbitrarily large density values in such cases and their training objective becomes ill-conditioned on data distributions that lie on manifolds. This well-known problem~\cite{gemici2016normalizing,brehmer2020flows} is typically tackled by artificially increasing the dimensionality of the data distribution. Practically, this is achieved by adding noise to the data~\cite{kingma2018glow,yang2019pointflow,klokov2020discrete}. While this makes the optimization of \acp{nf} feasible, it produces the undesirable side effect of reduced quality of generated data. Most prior works tackling this problem make strong assumptions about the underlying manifold~\cite{gemici2016normalizing,cunningham2020normalizing,brehmer2020flows,mathieu2020riemannian,horvat2021denoising} - i.e.\ regarding its structure or dimensionality. However, such assumptions are typically impractical in the case of real-world applications. Notably, SoftFlow~\cite{kim2020softflow} proposed a practical framework for learning distributions on lower dimensional manifolds using \acp{nf} and evaluated its benefit on 3D point clouds.

This work approaches this problem from a novel perspective. Given a \ac{nf} trained on a data distribution with added noise, we generate samples from the original distribution which resides on the lower dimensional manifold. We achieve this by recognizing that the regions of maximum likelihood locally present an implicit representation of the manifold if the added noise is sufficiently small. Using this observation, we propose an optimization objective that recovers the most likely point on the manifold, given a sample from the distribution with added noise. We show that our approach consistently improves performance when modeling 3D point clouds as well as images. Furthermore, we leverage the benefits of an explicit generative model to derive an approach that improves our method and reduces the overhead of test time optimization for the specific case of 3D point clouds. To this end, we use the normal direction extracted from the gradient of the log-likelihood and log-likelihood itself to perform Poisson surface reconstruction~\cite{kazhdan2006poisson} to efficiently refine the generated 3D point clouds.

%% file: sections/related_work.tex
\subsection{Modeling 3D Point Clouds}

Latent GAN~\cite{achlioptas2018learning} models the latent space of a pretrained autoencoder using a Generative Aversarial Network~\cite{goodfellow2014generative}. AtlasNet~\cite{groueix2018papier} learns to map 2D patches onto surfaces in 3D by minimizing the Chamfer distance. They also propose to post-process reconstructions by performing Poisson surface reconstruction. Therefore they densely sample points from their reconstructed mesh and generate oriented normals by shooting rays from infinity on the point cloud. In contrast, our approach yields oriented normals more efficiently with fewer points, since the direction perpendicular to the surface is readily available as the gradient of the log-likelihood of the \ac{nf}. ShapeGF~\cite{cai2020learning} models 3D point clouds using score matching~\cite{hyvarinen2005estimation}. Therefore, they train a \ac{nn} to estimate the gradient of the log-likelihood of the distribution of points. At inference time, they perform temperature annealed Langevin dynamics~\cite{welling2011bayesian,song2019generative} to transform a random set of points into realistic 3D shapes. Recently, \acp{dpm}~\cite{ho2020denoising} have shown strong performance on point clouds~\cite{luo2021diffusion}. Similar to ShapeGF, they transform an initial set of random points into realistic 3D point clouds. However, instead of relying on an unnormalized density function \acp{dpm} learn to approximately invert a diffusion process. Similar to \acp{dpm} and ShapeGF, our method also post-processes an initial estimate of the point cloud by maximizing the log-likelihood of the points. However, we require fewer update steps compared to ShapeGF, since our initial estimate of the point cloud is more accurate. Moreover, our method has the advantage of an explicit generative model. For instance, this is useful when attempting anomaly detection or when post-processing the generated point cloud with Poisson surface reconstruction (\secref{subsection:sampling_manifold_poisson}). 

Moreover, \acp{nf} have been applied to 3D point clouds. In a pioneering work, PointFlow~\cite{yang2019pointflow} demonstrated promising performance by applying continuous \acp{nf}. Later, Discrete PointFlow (DPF)~\cite{klokov2020discrete} improved upon PointFlow by using discrete \acp{nf}, namely RealNVPs~\cite{dinh2016density}, which are computationally more efficient while providing better performance. Concurrently, C-Flow~\cite{pumarola2020c} proposed a novel conditioning scheme for conditional \acp{nf} which they applied to point clouds. Further, the authors of \cite{postels2021go} and \cite{kimura2021chartpointflow} proposed simultaneously to model point clouds using a mixture of \acp{nf} which lead to better reconstruction performance.

\subsection{Normalizing Flows on Manifolds}

Real world data often resides on lower dimensional manifolds embedded in high dimensional space. This is a well-known problem when applying \acp{nf} to model realistic data distributions~\cite{gemici2016normalizing,cunningham2020normalizing,brehmer2020flows,mathieu2020riemannian,kim2020softflow,horvat2021denoising}. In~\cite{gemici2016normalizing}, the authors propose an approach to explicitly learn \acp{nf} on spheres in n-dimensional space. More recently, Brehmer and Cranmer~\cite{brehmer2020flows} proposed an alternating optimization approach that simultaneously learns the data distribution and the underlying manifold. Furthermore, \cite{cunningham2020normalizing} propose an end-to-end trainable approach to learning distributions on manifolds. Horvat and Pfister~\cite{horvat2021denoising} add noise to the data distribution and explicitly split the dimensions of a \ac{nf} into two sets, where one is assumed independent of the noise. Overall, most methods make explicit assumptions about the underlying structure and dimensionality of the manifold. In contrast, this work does not make any assumptions about the dimensionality of the manifold and only requires the manifold to be sufficiently smooth compared to the magnitude used to inflate the data distribution. Most relevant to this work is the recently proposed SoftFlow~\cite{kim2020softflow}. The authors model manifolds using \acp{nf} by training a \ac{nf} conditioned on the standard deviation $\sigma$ of the Gaussian noise added to the data distribution. During inference they set $\sigma = 0$ to sample from the original manifold. This work shows that one can achieve similar or better performance without training a model on all possible noise magnitudes. We achieve this by recognizing that a \ac{nf} trained on noisy data implicitly represents the manifold in regions of maximum likelihood.

%% file: sections/method.tex

This section presents a new approach to sample from distributions that reside on manifolds using \acp{nf}. One can regard those as delta distributions in a higher dimensional space. We first revisit \acp{nf} and discuss their challenges (\secref{subsection:nfs_manifolds}). We then demonstrate how \acp{nf} implicitly represent manifolds and introduce a general framework for sampling from this implicit representation (\secref{subsection:sampling_manifold})). Finally, we propose an efficient algorithm based on Poisson surface reconstruction~\cite{cai2020learning} for 3D point clouds (\secref{subsection:sampling_manifold_poisson}).

Therefore, let $\Omega$ denote a m-dimensional compact manifold embedded in n-dimensional Euclidean space, $\Omega \in \mathbb{R}^n$ where $m<n$. Capital letters $X$ refer to random variables with their corresponding probability density $P_X$ and lower case letters refer to their realizations $x\sim P_X(x)$.

\subsection{The Curse of Invertibility - Normalizing Flows on Manifolds}\label{subsection:nfs_manifolds}

\acp{nf} are a family of flexible parametric models for estimating probability distributions based on invertible \acp{nn}. Consider an invertible transformation $F_{\theta}: \mathbb{R}^n \rightarrow \mathbb{R}^n$ with parameters $\theta$ - parameterized by a \ac{nn}. In order to learn a probability distribution $P_X(x)$ with $x \in \mathbb{R}^n$, \acp{nf} parameterize the latter using the change of variable formula 
\begin{equation*}
    P_X(x) = P_Y(F_{\theta}^{-1}(x))\textrm{det}\left| J_{F_{\theta}^{-1}}(x) \right|
\end{equation*}
where $J_{F_{\theta}^{-1}}(x)$ is the Jacobian of $F_{\theta}^{-1}$ at $x$ and $Y$ is a random variable with known parametric distribution $P_Y$. The parameters $\theta$ are learned by maximizing the log-likelihood:
\begin{align}\label{eq:nll_flows}
\begin{split}
    \log(P_X(x)) &= \log(P_Y(F_{\theta}^{-1}(x))) \\
    &+ \log\left(\textrm{det}\left| J_{F_{\theta}^{-1}}(x) \right|\right)
\end{split}
\end{align}
From \eref{eq:nll_flows} we observe the main challenge of implementing \acp{nf} is the design of invertible \ac{nn} layers allowing to evaluate the logarithm of their Jacobian determinant efficiently. 

Consider a random variable $S$ that exists on a m-dimensional manifold $\Omega$ which is distributed according to $P_S(x) = 0$ $\forall x \notin \Omega$. Assuming $P_Y(x)$ has full support in $\mathbb{R}^n$, \eref{eq:nll_flows} yields an ill-posed optimization problem since $\log\left(\textrm{det}\left| J_{F_{\theta}^{-1}}(x) \right|\right)$ becomes infinitely large when squeezing $\mathbb{R}^n$ in its entirety onto $\Omega$. Likewise, choosing $P_Y(x)$ without full support in $\mathbb{R}^n$ renders the optimization problem equally ill-posed. The initial $F_{\theta}^{-1}$ is required to be an injective transformation from $\Omega$ and the support of $Y$. Otherwise the first term in \eref{eq:nll_flows}, $\log(P_Y(F_{\theta}^{-1}(x)))$, would tend to $-\infty$. This can also be observed in the toy example depicted in \figref{fig:toy_example} where the vanilla \ac{nf} is not able to successfully learn the distribution which lies on the 1-sphere. 

In practice these optimization difficulties are bypassed by adding noise to $S$~\cite{kingma2018glow,yang2019pointflow,klokov2020discrete,kim2020softflow}. We refer to this as inflating the distribution $P_S(x)$.  Formally, this corresponds to convolving $P_S$ with a noise distribution $P_N$:
\begin{equation}\label{eq:nll_convolution}
    P_X(x) = \int_{\mathbb{R}^n} P_N(x-s)P_S(s) ds
\end{equation}
A common choice for $P_N$ is the Gaussian distribution. Learning the inflated distribution $P_X$ instead of $P_S$ stabilizes training with \eref{eq:nll_flows}. However, it inevitably leads to a loss of information about $S$ since the learned distribution $P_X$ deviates from the distribution of interest $P_S$. 

\subsection{Sampling from the Implicit Manifold Representation}\label{subsection:sampling_manifold}

The goal of this work is to generate samples from $P_S$ given knowledge of $P_X$ and $P_N$. We achieve this by choosing the form of $P_N$ such that it allows us to generate samples from $P_S$ on $S$ given samples from $P_X$. We note that this is an ill-posed problem for general $P_N$. We first assume that $S$ is uniformly distributed on $\Omega$. Further, we ensure that $P_N$ is unimodal and zero-centered. We achieve this by construction, e.g. by choosing $P_N=\mathcal{N}(x;0,\Sigma)$. \textit{This allows us to identify the maximum of $P_X$ as an implicit representation of $S$}. Subsequently, we here set $P_N$ to be a n-dimensional Gaussian - $P_N(x) = \mathcal{N}(x;0,\Sigma)$ with $\Sigma=\sigma^2 \textrm{I}$. Given this insight, our primary concern is subsequently to develop an approach to project a sample $x \sim P_{X}(x)$ back to $S$.

Naturally, the quality of this representation deteriorates for larger noise magnitudes $\sigma$. In order to represent $S$ well, the noise magnitude $\sigma$ needs to be sufficiently small. In particular, consider the typical length scale $l$ of the manifold $S$. We define $l$ as the maximum side length of the m-dimensional hypercube that approximates $S$ well everywhere locally. In this case, the gradient of the log-likelihood only has a significant component that is perpendicular to $\Omega$ since $P_S$ is uniformly distributed on $\Omega$. This allows us to decode $S$ locally for a given point sampled from $P_X$ by following its gradient of the log-likelihood to its maximum.

Thus, assuming a uniformly distributed $S$ and that $F_{\theta}$ sufficiently well estimates $P_X$, we can sample $s\sim P_S(x)$ on $S$ by: 1.) drawing a sample $x'\sim P_X(x)$ and 2.) solving the following multi-objective optimization problem that equates to finding the maximum of $P_X(x)$ closest to $x$:
\begin{equation}\label{eq:optimization_criterion}
\begin{aligned}
    s &=\underset{x\in \mathbb{R}^n}{\textrm{argmin}} \left( ||x - x'||_2^2 -\log(P_X(x)) \right)
\end{aligned}
\end{equation}
Note that even uniformly distributed $P_S$ are practically relevant as they are often assumed for point clouds of 3D shapes. However, it is also interesting to consider the case of non-uniformly distributed $P_S$. For such $P_S$, applying \eref{eq:optimization_criterion} would still generate samples that lie on $\Omega$. However, the resulting distribution may not resemble $P_S$ since for non-uniform $P_S$ we do not have $\nabla P_X(x) = 0$ everywhere on $S$. Nevertheless, for sufficiently smooth $P_S$ and small $\sigma$ \eref{eq:optimization_criterion} still yields a good estimate of $P_S$. Specifically, we consider $\sigma$ such that $\sigma \ll \frac{1}{\underset{s\in \Omega}{\textrm{max}} \left( \left| \nabla_{\Omega} P_S(x) \right| \right)}$, where $\nabla_{\Omega}$ denotes the derivative along $\Omega$. In this case we can still approximate $P_S$ locally as a uniform distribution on an m-dimensional hypercube and, thus, obtain samples from $P_S$.

We optimize \eref{eq:optimization_criterion} by minimizing the objective:
\begin{equation}\label{eq:optimization_loss}
\begin{aligned}
  \mathcal{L}(x) = -\log(P_X(x)) + \lambda ||x - x'||_2^2
\end{aligned}
\end{equation}
$\lambda$ denotes a hyperparameter that is set by optimizing performance on the validation set. Interestingly, $\frac{\textrm{d}\lambda\|x - x'\|_2^2}{\textrm{d}x} = -\frac{\textrm{d}\log\left( \mathcal{N}(x;x',\frac{1}{\lambda}I) \right)}{\textrm{d}x} + C$ with $C$ constant in $x$. This is the kernel used to inflate $P_S$. Thus, from a Bayesian perspective we can also interpret it as the negative logarithm of a prior distribution on the location of the most likely point $s$ on $\Omega$ that generated $x'$. In this light, $P_X(x)$ resembles the likelihood function learned from the data and we understand the optimum of \eref{eq:optimization_loss} as the MAP estimate of $s$ given $x'$. Subsequently, we refer to Log-Likelihood Maximization according to \eref{eq:optimization_criterion} as LLM.

\textbf{Choice of $\lambda$.} We implement $\lambda$ as a global hyperparameter which is optimized using the validation set. However, a locally varying $\lambda = \lambda(x)$ may improve upon this restrictive assumption. While we find empirically that a global $\lambda$ already yields significant advantages, we consider a locally varying $\lambda$ as a promising future research avenue.

\subsection{An Adjusted Algorithm for 3D Point Clouds}\label{subsection:sampling_manifold_poisson}


While \eref{eq:optimization_loss} gives us a practical and general framework for generating samples from $P_S$ using $P_X$, it can be computationally demanding since it requires multiple gradient updates. This is particularly challenging when using large neural networks to parameterize $P_X(x)$ as it is often the case for real-world data distributions such as images~\cite{kingma2018glow,grathwohl2018ffjord,behrmann2019invertible} and 3D point clouds~\cite{yang2019pointflow,klokov2020discrete,postels2021go}. 

We here view 3D point clouds as distributions on 2D manifolds in $\mathbb{R}^3$ and derive an approach that circumvents multiple gradient updates for point clouds. In contrast to other data modalities such as images, generating a 3D point cloud requires sampling a large set of points which accurately represents the entire distribution at once. Thus, we can generate samples on the manifold given global information about the entire set of points. Hereafter, we focus on the case where $\Omega$ refers to a 2D manifold embedded in $\mathbb{R}^3$ and $P_S$ refers to a uniform distribution on $\Omega$.

We consider the gradient of the log-likelihood parameterized by the \ac{nf} $\nabla\log(P_X(x))$. Given sufficiently small $\sigma$, $\nabla\log(P_X(x))$ is parallel to a surface normal on $\Omega$ for uniform $P_S$ in the vicinity of $\Omega$. We illustrate this in \figref{fig:toy_example}. Such normal vectors can be used to fuel off-the-shelf surface reconstruction algorithms such as Poisson surface reconstruction~\cite{kazhdan2006poisson}. Poisson surface reconstruction uses point sets and oriented surface normals to construct an implicit function of a watertight 3D shape. The surface can then be reconstructed by locating the iso-surface of the implicit function. While we here work with Poisson surface reconstruction, our approach naturally transfers to any surface reconstruction algorithm that relies on surface normals~\cite{berger2017survey}. Subsequently, we term this this approach LL-Poisson.

The proposed algorithm is depicted in \aref{alg:poisson_reconstruction}. Given a set of K\textsubscript{1} points $\mathcal{X}$ sampled from $P_X$, we apply the following routine. We first compute the gradient $\nabla\log(P_X(x))$ at each point and normalize it to unit length. Then, we create consistent normals by propagating the orientation of these vectors using a graph created by considering the k nearest neighbors of each point. These K vector-point pairs are then used to obtain a mesh representing $\Omega$. We refine this mesh using $\log(P_X(x))$ by removing vertices that have lower likelihood than the $\alpha$-percentile of the original set of points $\mathcal{X}$. The latter is particularly crucial for removing vertices that reside far away from $\Omega$ and when dealing with non-watertight 3D shapes since Poisson surface reconstruction generates watertight meshes. Finally, we uniformly sample K\textsubscript{2} points from the resulting mesh. 

\begin{algorithm}[t!]
\caption{Generating Samples on $\Omega$ }\label{alg:poisson_reconstruction}
\begin{algorithmic}
\Require $F_{\theta}$, initial point set $\mathcal{X}$, k, K\textsubscript{2}, $\alpha$
\State $\textrm{G} \gets \nabla\log(P_X(\mathcal{X}))$
\State $\textrm{G} \gets \textrm{normalize}(\textrm{G})$ \Comment{Normalized to unit length}
\State $\textrm{G} \gets \textrm{propagate\_orientation}(\textrm{G}, \textrm{k})$ \Comment{Propagates orientation according to k nearest neighbors}
\State $\textrm{mesh} \gets \textrm{poisson\_reconstruction}(\mathcal{X}, \textrm{G})$
\State $\textrm{perc}_{\alpha} \gets \textrm{percentile}(\alpha, P_X(\mathcal{X}))$
\State $\textrm{mesh} \gets \textrm{remove\_vertices}(\textrm{mesh}, \textrm{perc}_{\alpha})$ \Comment{Remove unlikely vertices}
\State $\mathcal{X}_{\Omega} \gets \textrm{sample\_uniformly}(\textrm{mesh}, \textrm{K\textsubscript{2}})$ \Comment{Generate K\textsubscript{2} uniform samples from the mesh}
\Return $\mathcal{X}_{\Omega}$
\end{algorithmic}
\end{algorithm}

%% file: sections/experiments.tex
We first present a toy example to foster an intuitive understanding of the effect of optimizing \eref{eq:optimization_criterion} (\secref{fig:toy_example}). Then, we compare ManiFlow based on log-likelihood maximization (LLM) and LL-Poisson with SoftFlow~\cite{kim2020softflow} (\secref{subsubsection:softflow}) and other point cloud representations (\secref{subsubsection:sota_autoencoding}) for point cloud autoencoding on ShapeNet~\cite{shapenet2015}. Finally, we demonstrate that ManiFlow scales to high-dimensional data distributions on image generation (\secref{subsection:image_modeling}). 

\subsection{Synthetic Data}\label{subsection:toy_example}

We perform experiments on artificial distributions in 2D. We consider two cases: a uniform and a non-uniform distribution on a 1-sphere. We train a RealNVP~\cite{dinh2016density} with 5 coupling layers for 5000 epochs on both distributions. More training details can be found in the supplement.

The distributions and results are depicted in \figref{fig:toy_example}. As expected we observe that a standard \ac{nf} directly trained on the clean data fails to generate samples that match the target distribution. We further verify that the problem can be mitigated by adding Gaussian noise to the data at training time. We find that performing post-training LLM according to \eref{eq:optimization_criterion} allows us to sample from the original manifold on the 1-sphere in accordance with the ground truth distribution. \figref{fig:toy_example} also visualizes the direction of the gradients of the log-likelihood. The gradients are approximately perpendicular to the 1-sphere in the vicinity of the manifold in case of the uniform as well as the non-uniform distribution. The direction only starts to deviate in regions of very low probability density. This observation underlines our discussion in \secref{subsection:sampling_manifold} and motivates us to use \acp{nf} in combination with Poisson surface reconstruction (see \secref{subsection:sampling_manifold_poisson}).

\subsection{3D Point Cloud Autoencoding}\label{subsection:pc_autoencoding}

\input{figs/shapenet/vis}

3D point clouds are a widespread data modality. Since they are often exclusively comprised of surface points, they denote an important testbed for methods that generate data on lower dimensional manifolds. We evaluate on point cloud autoencoding since we are primarily interested in the ability of ManiFlow to represent 3D point clouds.

\noindent\textbf{Dataset.} We evaluate on the ShapeNet dataset~\cite{shapenet2015}. We use the categories: airplanes, cars, and chairs. We train on the presampled point clouds provided by PointFlow~\cite{yang2019pointflow}, which are also used by SoftFlow~\cite{kim2020softflow}, DPM~\cite{luo2021diffusion} and ShapeGF~\cite{cai2020learning}. Each shape consists of 15000 points which we randomly subsample to 2048 points during training and test time. Prior to subsampling each shape is individually normalized such that it has zero mean and unit variance. 

\noindent\textbf{Evaluation.} At test time we decode 2048 points from the trained models and compare them to 2048 points uniformly sampled from the ground truth. We repeat each evaluation 5 times and report the mean result. We use three metrics for measuring the quality of reconstructed point clouds. We follow prior work and compute the \ac{cd} and \ac{emd}~\cite{achlioptas2018learning}. If not stated otherwise, we follow prior work~\cite{yang2019pointflow} and report results on CD multiplied by $10^4$ and results on EMD by $10^2$. We further report the F1-score (F1)~\cite{knapitsch2017tanks}. We use a threshold of $\tau=10^{-4}$ for the F1-score. We also report the performance of an oracle as an upper bound to the performance. The oracle performance is computed by comparing two independently sampled subsets of the ground truth point cloud.

\noindent\textbf{Encoder.} In an autoencoder an encoder model computes a latent representation of a point cloud that, subsequently, a decoder model uses to reconstruct the original point cloud. We follow PointFlow~\cite{yang2019pointflow} and parameterize the encoder with a PointNet~\cite{qi2017pointnet}, which produces permutation-invariant latent representations. Specifically, we use convolutions with filter sizes of 128, 256 and 512 followed by a max pooling layer. The resulting 512-dimensional representation is transformed into a 256-dimensional latent representation using a single fully-connected layer. As a decoder model we use a conditional \ac{nf}. In particular, when comparing with SoftFlow~\cite{kim2020softflow}, we follow their experimental setup and use a \ac{nf} based on autoregressive layers and invertible $1\times1$ convolutions (\secref{subsubsection:softflow}). When comparing against methods for modeling point clouds (\secref{subsubsection:sota_autoencoding}), we adopt a RealNVP similar to the one used in DPF~\cite{klokov2020discrete}. The RealNVP consists of 64 coupling layers where each coupling layer consists of two fully-connected layers with a hidden dimensionality of 64. The first fully-connected layer is followed by a Batch Normalization layer and the swish activation function~\cite{ramachandran2017searching}. Further details are in the supplement.

\input{tables/shapenet_softflow}

\noindent\textbf{Optimization.} In \secref{subsubsection:softflow} we precisely follow the setup used by SoftFlow~\cite{kim2020softflow}. We only deviate from it when training a basic \ac{nf} in absence of the SoftFlow framework. In this case we simply use Gaussian noise of a fixed magnitude during training instead of randomly sampling the noise variance from a range of values at each iteration. We set the standard deviation of the noise to $0.02$. In \secref{subsubsection:sota_autoencoding} we train the autoencoder for 1500 epochs using a batch size of 128 using the Adam optimizer~\cite{kingma2014adam}. We set the initial learning rate to $6.4\times10^{-4}$ and divide it by 4 after 1200 and 1400 steps. We exponentially decay the standard deviation of the Gaussian noise added to the data from 0.25 to 0.02 between the epochs 100 and 1200. Each model is optimized on 4 V100 GPUs. For more details we refer to the supplement.

\noindent\textbf{ManiFlow - LLM.} When performing LLM (\eref{eq:optimization_criterion}) we use the Adam optimizer and a fixed learning rate of $10^{-3}$. We found 25 gradient updates to be sufficient. We choose a regularization parameter $\lambda$ in \eqref{eq:optimization_loss} such that performance on validation set is optimized. Note that $\lambda$ is sensitive to the expressiveness of the \ac{nf}. More expressive \ac{nf} lead to larger log-likelihoods, which requires larger $\lambda$. In \secref{subsubsection:softflow} we use $\lambda=2.0$ and in \secref{subsubsection:sota_autoencoding} we use $\lambda=4000$.

\noindent\textbf{ManiFlow - LL-Poisson.} We first create an initial set of K\textsubscript{1}=10000 points using the \ac{nf} and compute the gradient of the log-likelihood for each of them. Then, we ensure consistent orientation of the resulting vectors by propagating their orientation based on $k=20$ nearest neighbors. We perform Poisson surface reconstruction using maximum depth of 9 provided by Open3D~\cite{Zhou2018}. Finally, we set $\alpha=0.05$ and sample K\textsubscript{1}=2048 points uniformly from the resulting mesh.

\subsubsection{Comparison with SoftFlow}\label{subsubsection:softflow}

\input{tables/shapenet_dpm}

Here we compare ManiFlow with SoftFlow~\cite{kim2020softflow}. We report autoencoding performance of \ac{nf} with and without the SoftFlow framework and apply ManiFlow based on LLM/LL-Poisson to the \ac{nf} in absence of SoftFlow. We also perform LLM on the \ac{nf} trained with SoftFlow. 

The results are in \tabref{table:softflow_comparison}. We make several observations. ManiFlow LLM/LL-Poisson consistently improves reconstruction performance. Secondly, despite the absence of the SoftFlow framework at training time, a vanilla \ac{nf} is able to outperform a \ac{nf} trained with SoftFlow when applying ManiFlow. Further, the performance of SoftFlow can be reliably improved when applying LLM to it, indicating that SoftFlow can be enhanced by applying ManiFlow. Lastly, we note that LL-Poisson tends to outperform LLM despite of being twice as fast. This demonstrates that refining the entire point set as a whole based on the \ac{nf}'s log-likelihood is superior over projecting each point individually.

\subsubsection{Further Comparison on Autoencoding.}\label{subsubsection:sota_autoencoding}

\input{tables/poisson_ablation}

We compare ManiFlow with recent methods proposed for representing point clouds - AtlasNet~\cite{groueix2018papier}, PointFlow~\cite{yang2019pointflow}, ShapeGF~\cite{cai2020learning} and DPM~\cite{luo2021diffusion}. The results are in \tabref{table:full_shapenet_comparison}. ManiFlow LLM/LL-Poisson outperforms standard \acp{nf}, which is consistent with previous results. Further, \acp{nf} equipped with ManiFlow are able to clearly outperform AtlasNet~\cite{groueix2018papier} which was not possible before~\cite{klokov2020discrete,postels2021go} while allowing likelihood evaluation. While recent implicit generative models, such DPM and ShapeGF, demonstrate strong performance, ManiFlow is able to reduce the gap. Moreover, \figref{fig:shapenet_rendering} visualizes reconstructed point clouds from DPM and ManiFlow. LL-Poisson tends to create smooth surfaces.

For ManiFlow LL-Poisson, it is important to use large number of points K\textsubscript{1} for obtaining competitive performance. To verify that this does not equally help other models, we also post-process point clouds generated by DPM using Poisson surface reconstruction. We apply the same hyperparameters as for ManiFlow with LL-Poisson. The only difference is the inability to evaluate the likelihood of each point. Therefore, we estimate normals based on Principal Components Analysis (PCA). We report the result in \tabref{table:dpm_pp}. Although using a large number of points for creating the mesh, the final autoencoding performance deteriorates in absence of the ability to evaluate the log-likelihood.

Similarly, we investigate the impact of using the log-likelihood provided by the \ac{nf} when using LL-Poisson. Therefore, we evaluate our model using LL-Poisson and with Poisson surface reconstruction with normals estimated via PCA on the airplane category of ShapeNet. The results are reported in \tabref{table:maniflow_poisson_ablation}. LL-Poisson outperforms the baseline on every metric. We conclude that the log-likelihood is an important ingredient for LL-Poisson.

Given the importance of the information stored in the log-likelihood, we are interested in determining whether LL-Poisson is beneficial when reconstructing a mesh from a sparse point cloud. Therefore, we set K\textsubscript{1}=256 and perform again surface reconstruction with and without information from the log-likelihood. We expect that for a good mesh uniformly sampled points from it demonstrate good autoencoding performance. We show the results in \tabref{table:maniflow_poisson_sparse_pointclouds}. We observe an even larger performance improvement when using LL-Poisson over the baseline in the case of sparse point clouds. We conclude that LL-Poisson can be useful when reconstructing meshes from sparse point clouds.

\input{tables/sparse_point_clouds_and_fid_images_gaussian}

\subsection{Generative Modelling on Images}\label{subsection:image_modeling}

So far we have conducted experiments low-dimensional data. In the following we further evaluate ManiFlow based on LLM on image generation to investigate whether it scales to high-dimensional data distributions. To this end, we train a GLOW~\cite{kingma2018glow} on MNIST~\cite{lecun1998gradient}, CIFAR10~\cite{krizhevsky2009learning} and CelebA~\cite{liu2015faceattributes}. We train each model for 400 epochs using a batch size of 256 on MNIST/CIFAR10 and 128 on CelebA. We use the Adam~\cite{kingma2014adam} optimizer and a fixed learning rate of $10^{-4}$. When training \acp{nf} on images one typically uses additive uniform noise~\cite{kingma2018glow}. Since uniform noise violates our assumption of unimodal noise, we replace it with additive Gaussian noise. We choose the standard deviation of the Gaussian noise such that it has the same entropy as the uniform distribution used in the original work, i.e.\ $\sigma=0.00756$. After training we perform 50 steps of LLM using $\lambda=0.2$ and a learning rate of $10^{-3}$. We report the Fr\'echet Inception Distance (FID)~\cite{heusel2017gans} which is widely used to quantify the perceptual quality of generated images. 

\tabref{table:images_fid} shows quantitative results on image generation. We observe a reliable and clear improvement of the FID-score when applying ManiFlow. Thus, ManiFlow paired with LLM based on \eref{eq:optimization_criterion} scales to high-dimensional data.

%% file: figs/shapenet/vis.tex
\begin{figure*}[h!]
\begin{center}

    \renewcommand{\figsize}{0.15}
    \newcommand{\planesize}{4}
    \newcommand{\carsize}{3.5}
    \newcommand{\chairsize}{2.2}
    
    \begin{tabular}{ c c c c c }
    \setlength{\tabcolsep}{0pt} 
    \renewcommand{\arraystretch}{0} 
    
        \centering 
    
      GT  & DPM~\cite{luo2021diffusion} & NF & NF\textsubscript{LLM} & NF\textsubscript{LL-Poisson} \\
      
      \includegraphics[width=\figsize\linewidth,trim={\planesize cm \planesize cm \planesize cm \planesize cm},clip]{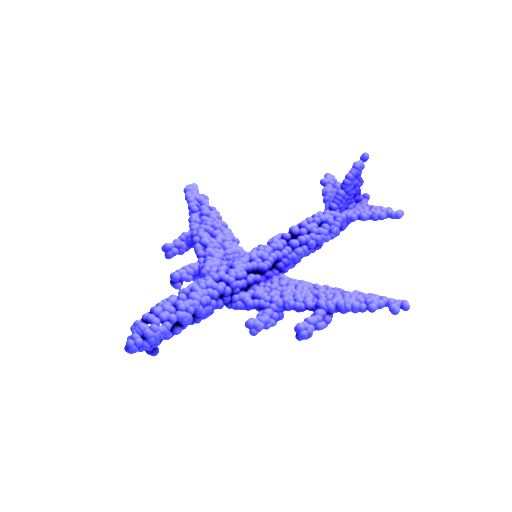} &
      \includegraphics[width=\figsize\linewidth,trim={\planesize cm \planesize cm \planesize cm \planesize cm},clip]{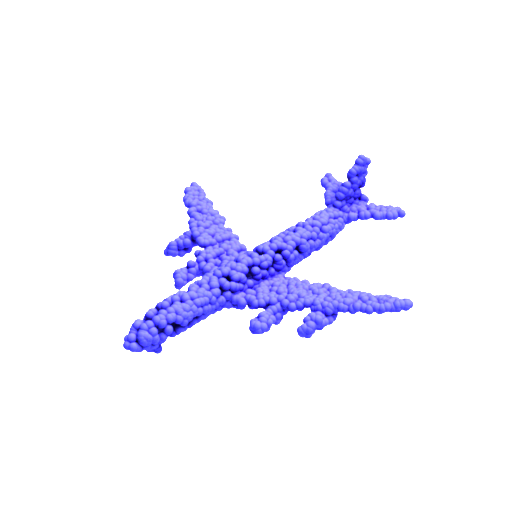} &
      \includegraphics[width=\figsize\linewidth,trim={\planesize cm \planesize cm \planesize cm \planesize cm},clip]{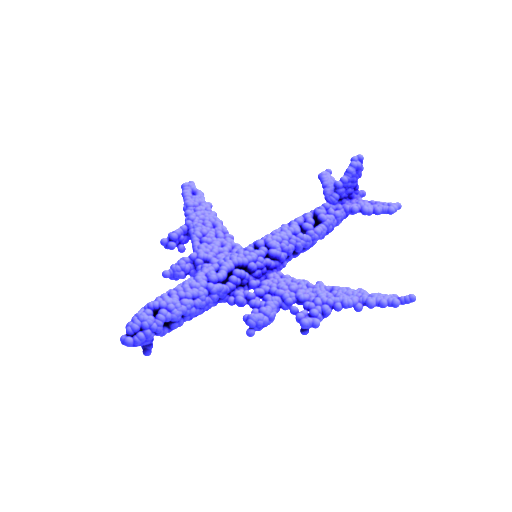} &
      \includegraphics[width=\figsize\linewidth,trim={\planesize cm \planesize cm \planesize cm 4cm},clip]{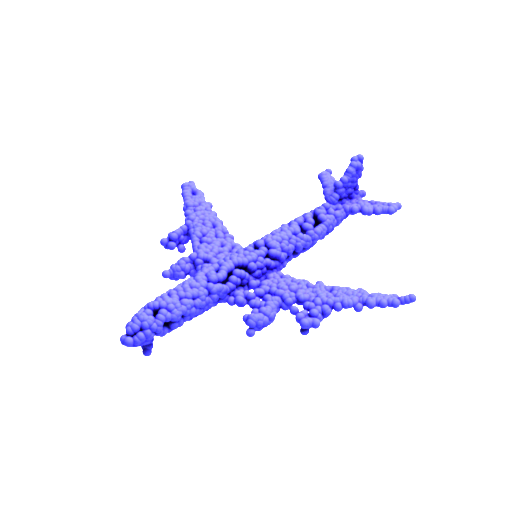} &
      \includegraphics[width=\figsize\linewidth,trim={\planesize cm \planesize cm \planesize cm \planesize cm},clip]{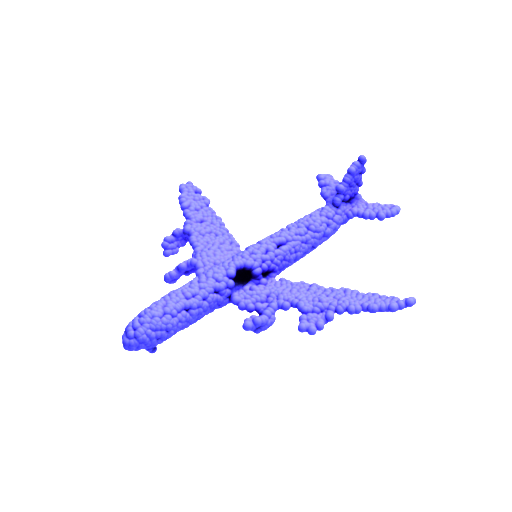} \\
      
      \includegraphics[width=\figsize\linewidth,trim={\carsize cm \carsize cm \carsize cm \carsize cm},clip]{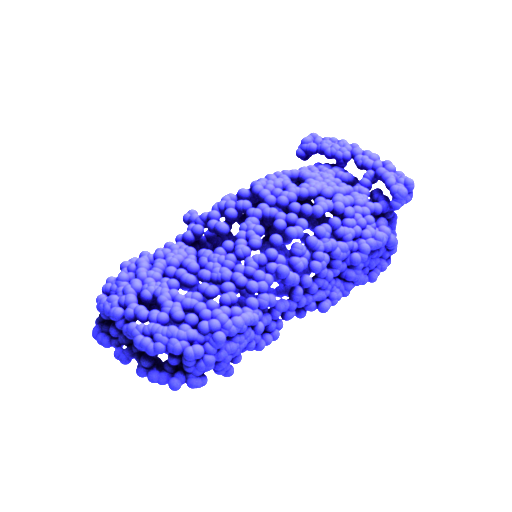} &
      \includegraphics[width=\figsize\linewidth,trim={\carsize cm \carsize cm \carsize cm \carsize cm},clip]{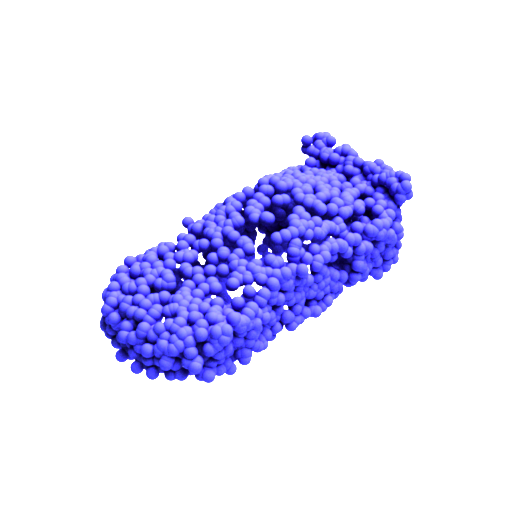} &
      \includegraphics[width=\figsize\linewidth,trim={\carsize cm \carsize cm \carsize cm \carsize cm},clip]{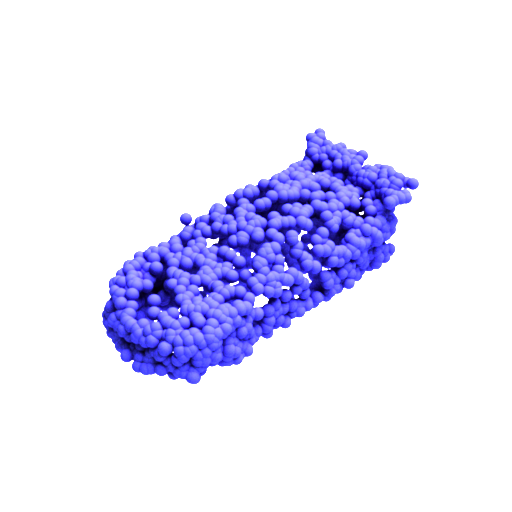} &
      \includegraphics[width=\figsize\linewidth,trim={\carsize cm \carsize cm \carsize cm \carsize cm},clip]{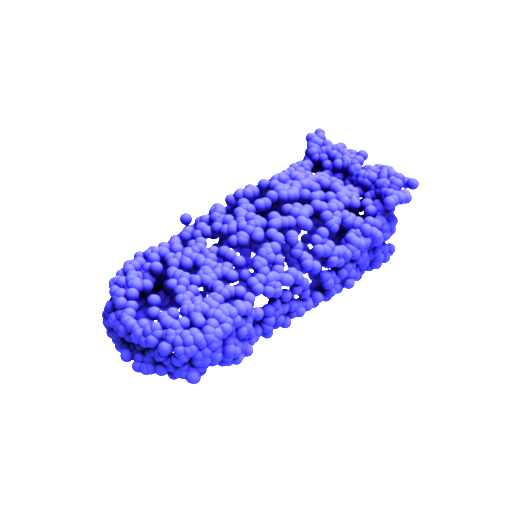} &
      \includegraphics[width=\figsize\linewidth,trim={\carsize cm \carsize cm \carsize cm \carsize cm},clip]{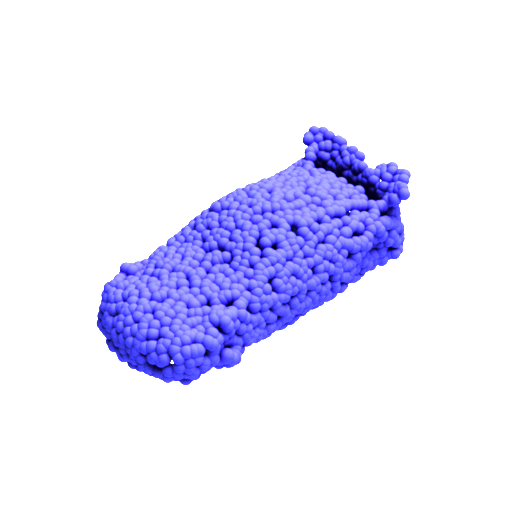} \\
      
      \includegraphics[width=\figsize\linewidth,trim={\chairsize cm \chairsize cm \chairsize cm \chairsize cm},clip]{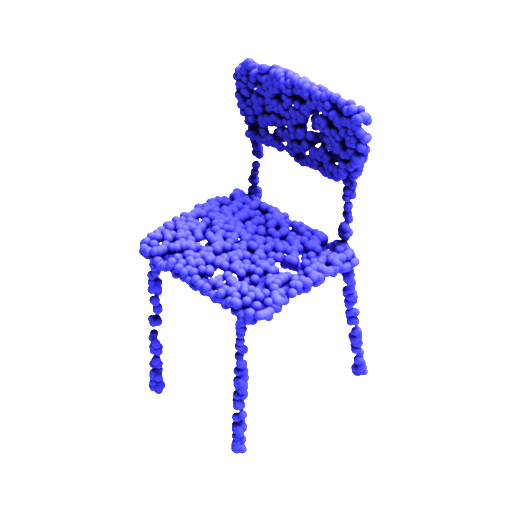} &
      \includegraphics[width=\figsize\linewidth,trim={\chairsize cm \chairsize cm \chairsize cm \chairsize cm},clip]{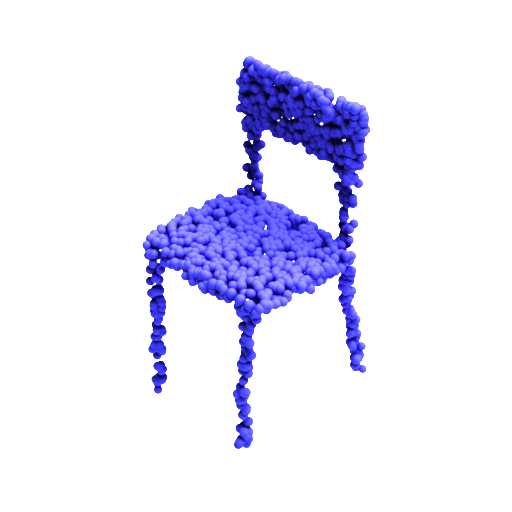} &
      \includegraphics[width=\figsize\linewidth,trim={\chairsize cm \chairsize cm \chairsize cm \chairsize cm},clip]{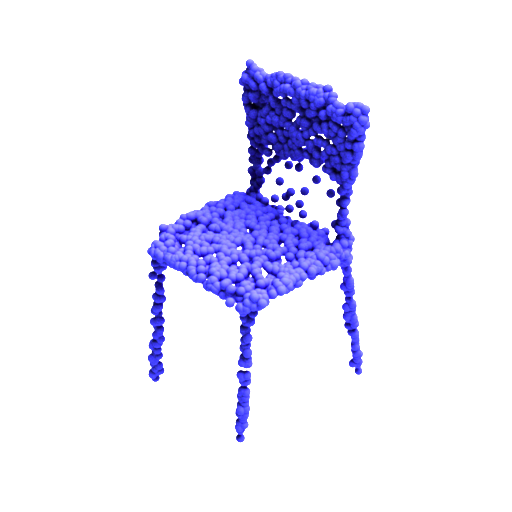} &
      \includegraphics[width=\figsize\linewidth,trim={\chairsize cm \chairsize cm \chairsize cm \chairsize cm},clip]{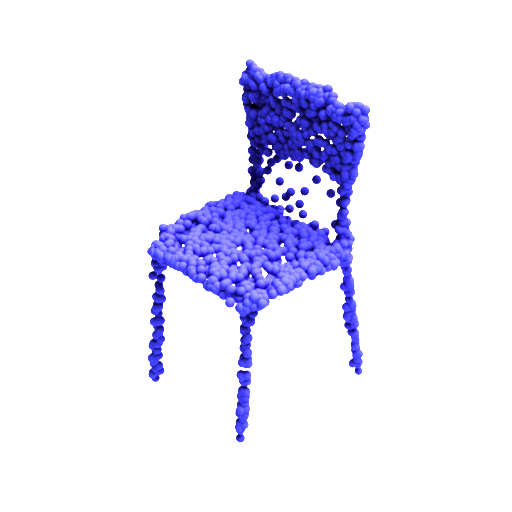} &
      \includegraphics[width=\figsize\linewidth,trim={\chairsize cm \chairsize cm \chairsize cm \chairsize cm},clip]{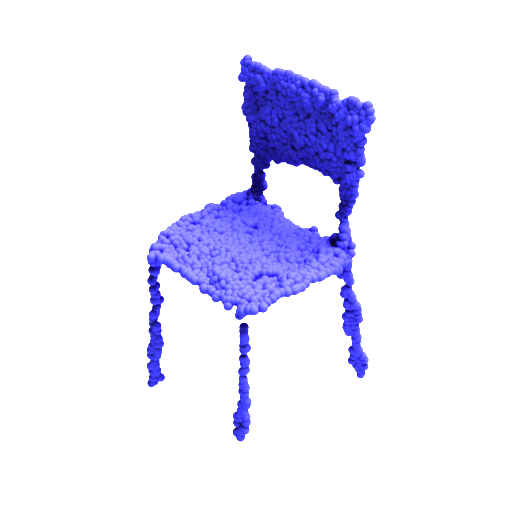} \\
      
    \end{tabular}
\end{center}
\caption{Qualitative comparison of the ground truth (GT), DPM~\cite{luo2021diffusion}, a RealNVP (NF) and a RealNVP with LLM/LL-Poisson (NF\textsubscript{LLM}/NF\textsubscript{LL-Poisson}). We observe that ManiFlow paired with LL-Poisson yields smoother surfaces.}\label{fig:shapenet_rendering}
\end{figure*}

%% file: tables/shapenet_softflow.tex
\begin{table}[t!]
\begin{center}
\begin{tabular}{|l|l|c|c|c|}
\hline
 & Method & CD$\downarrow$ & EMD$\downarrow$ & F1 $\uparrow$\\
\hline
\hline
\multirow{4}{*}{\rotatebox[origin=r]{90}{Airplane}}   
                            & NF         & 1.262             & 2.58              & 84.22             \\
                            & NF + SF        & 1.179             & 2.69              & 85.63             \\
                            & NF + SF + LLM   & 1.161             & 2.68              & 85.96             \\
                            & NF + LLM    & 1.151    & \textbf{2.56}     & 86.25    \\
                            & NF + LL-Poisson    & \textbf{1.132}            & 3.06     &  \textbf{87.26}   \\
\hline
\multirow{4}{*}{\rotatebox[origin=r]{90}{Car}}        & NF         & 6.978             & 5.18     & 22.36             \\
                            & NF + SF        & 6.876             & 5.23              & 23.19             \\
                            & NF + SF + LLM   & 6.850    & 5.22              & 23.44             \\
                            & NF + LLM    & 6.887             & 5.18     & \textbf{24.05}    \\
                            & NF + LL-Poisson    & \textbf{6.830}            & \textbf{5.01}     & 23.73    \\
\hline
\multirow{4}{*}{\rotatebox[origin=r]{90}{Chair}}      & NF         & 11.76             & 6.86              & 19.77             \\
                            & NF + SF        & 10.84             & \textbf{6.43}     & 20.86             \\
                            & NF + SF + LLM   & \textbf{10.83}    & \textbf{6.43}     & 21.13    \\
                            & NF + LLM    & 11.57             & 6.78              & 20.85             \\
                            & NF + LL-Poisson    & 11.60             & 6.69              & \textbf{21.91}             \\
\hline
\end{tabular}
\caption{Quantitative comparison of ManiFlow on point cloud autoencoding with SoftFlow~\cite{kim2020softflow} (SF) on the categories airplane, car and chair of ShapeNet. We train a vanilla \ac{nf} trained on noisy data with and without the SoftFlow framework. Post-training we apply either log-likelihood maximization (LLM) (\eref{eq:optimization_criterion}) or Poisson surface reconstruction (LL-Poisson). ManiFlow consistently outperforms SoftFlow and is even able to improve SoftFlow. CD is multiplied by $10^4$ and EMD by $10^4$.}
\label{table:softflow_comparison}
\vspace{-10mm}
\end{center}
\end{table}


%% file: tables/shapenet_dpm.tex
\begin{table}[t!]
\begin{center}
\begin{tabular}{|l|l|c|c|c|}
\hline
 & Method & CD$\downarrow$ & EMD$\downarrow$ & F1 $\uparrow$\\
\hline
\hline
\multirow{8}{*}{\rotatebox[origin=r]{90}{Airplane}}   & AtlasNet - Sphere ~\cite{groueix2018papier}        & 1.178             & 3.36              &  84.66            \\ 
                            & AtlasNet - Square ~\cite{groueix2018papier}        & 1.211   
                            & 3.42             &  84.30            \\
                            & PointFlow~\cite{yang2019pointflow}        & 1.213             & 2.76              &  85.31            \\
                            & ShapeGF~\cite{cai2020learning}        & 0.949             & 2.49              &  89.14            \\
                            & DPM~\cite{luo2021diffusion}        & \textbf{0.947}             & \textbf{2.19 }             &  \textbf{89.27 }           \\
                            & RealNVPs        & 1.139             & 2.77              &  85.69            \\
                            & RealNVPs + LLM        & 1.063             & 2.76              &  86.98            \\
                            & RealNVPs + Poisson      & 1.045             & 2.86              &  88.13            \\
                            \hline
                            & Oracle      &   0.722           &    1.95           &   92.31          \\
\hline\hline
\multirow{8}{*}{\rotatebox[origin=r]{90}{Car}}        & AtlasNet - Sphere ~\cite{groueix2018papier}        & 6.231             & 5.09              &  25.96            \\
                            & AtlasNet - Square ~\cite{groueix2018papier}        & 6.221   
                            & 5.19             &  25.79   \\
                            & PointFlow~\cite{yang2019pointflow}        & 6.540            & 5.16              &  24.02            \\
                            & ShapeGF~\cite{cai2020learning}        & 5.508             & 4.23              &  \textbf{28.91}            \\
                            & DPM~\cite{luo2021diffusion}        & \textbf{5.462}             & \textbf{3.96 }             &  28.77            \\
                            & RealNVPs        & 6.134             & 4.89              &  24.96            \\
                            & RealNVPs + LLM       & 5.976             & 4.90              &  26.99           \\
                            & RealNVPs + Poisson         & 5.928             & 4.30              &  26.44            \\
                            \hline
                            & Oracle      &     3.368         &   3.04            &   49.76          \\
\hline\hline
\multirow{8}{*}{\rotatebox[origin=r]{90}{Chair}}      & AtlasNet - Sphere ~\cite{groueix2018papier}        & 7.643             & 6.30             &  27.51            \\
                            & AtlasNet - Square ~\cite{groueix2018papier}        & 7.690   
                            & 6.74              &  27.13   \\
                            & PointFlow~\cite{yang2019pointflow}        & 10.094             & 6.46              &  21.33            \\
                            & ShapeGF~\cite{cai2020learning}        & 6.297             & 5.03              &  32.68            \\
                            & DPM~\cite{luo2021diffusion}        & \textbf{6.293}             & \textbf{4.27}              &  \textbf{32.59}           \\
                            & RealNVPs        & 8.071             & 5.31              &  25.59            \\
                            & RealNVPs + LLM      & 8.003             & 5.34              &  27.09            \\
                            & RealNVPs + Poisson         & 8.013             & 5.04              &  27.34           \\
                            \hline
                            & Oracle      &    2.762          &     3.07          &    56.02         \\
\hline
\end{tabular}
\caption{Quantitative comparison of autoencoding performance with AtlasNet~\cite{groueix2018papier}, PointFlow~\cite{yang2019pointflow}, ShapeGF~\cite{cai2020learning} and DPM~\cite{luo2021diffusion} on airplane/chair/car category of ShapeNet. ManiFlow reliably improves performance and enables \acp{nf} to outperform AtlasNet. Poisson slightly outperforms LLM. ManiFlow consistently outperforms SoftFlow and is even able to enhance the performance of SoftFlow. CD is multiplied by $10^4$ and EMD by $10^4$.}
\label{table:full_shapenet_comparison}
\vspace{-5mm}
\end{center}
\end{table}

%% file: tables/poisson_ablation.tex
\begin{table}[t!]
\centering
\centering
\begin{tabular}{|l|c|c|c|}
\hline
Poisson & CD$\downarrow$ & EMD$\downarrow$ & F1$\uparrow$\\
\hline
\hline
 \ding{55}        & \textbf{0.947}             & \textbf{2.19}              &  \textbf{89.27} \\
 \ding{51}        & 0.975             & 2.99              &  88.91 \\
\hline
\end{tabular}
\caption{Poisson surface reconstruction on point clouds generated by DPM~\cite{luo2021diffusion} on the airplane category. Unlike for \acp{nf}, we observe that the reconstruction quality degrades}
\label{table:dpm_pp}
\centering
\begin{tabular}{|l|c|c|c|}
\hline
Use LL & CD$\downarrow$ & EMD$\downarrow$ & F1$\uparrow$\\
\hline
\hline
 \ding{55}        & 1.077            & 3.01             &  87.50 \\
 \ding{51}        & \textbf{1.045}             & \textbf{2.86}              &  \textbf{88.13} \\
\hline
\end{tabular}
\caption{ManiFlow LL-Poisson with/without the log-likelihood (LL) of the \ac{nf} on the airplane category. The reconstruction quality degrades without LL of the \ac{nf}}
\label{table:maniflow_poisson_ablation}
\centering
\begin{tabular}{|l|c|c|c|}
\hline
Use LL & CD$\downarrow$ & EMD$\downarrow$ & F1$\uparrow$\\
\hline
\hline
 \ding{55}        & 24.256            & 7.73             &  41.53 \\
 \ding{51}        & \textbf{9.427}             & \textbf{6.91}              &  \textbf{54.83} \\
\hline
\end{tabular}
\caption{ManiFlow LL-Poisson with/without the log-likelihood (LL) of the \ac{nf} on sparse point clouds (256 points) of the the airplane category. The reconstruction quality degrades when not relying on the LL of the \ac{nf} more significantly on sparse point clouds}
\label{table:maniflow_poisson_sparse_pointclouds}
\end{table}

%% file: tables/sparse_point_clouds_and_fid_images_gaussian.tex
\begin{table}[t!]
\centering
\begin{tabular}{|l|c|c|c|}
\hline
LLM & MNIST & CIFAR10 & CelebA\\
\hline
\hline
 \ding{55}  & 30.0        &     72.7  & 51.7 \\
 \ding{51}  & \textbf{21.7}        &    \textbf{64.7}         & \textbf{43.9} \\
\hline
\end{tabular}
\caption{Quantitative evaluation of LLM on image generation. We report the FID score when training a GLOW model~\cite{kingma2018glow} on MNIST/CIFAR10/CelebA with/without LLM. During training Gaussian noise is added to the images. LLM improves generation performance}
\label{table:images_fid}
\end{table}

%% file: sections/conclusion.tex
This work introduced ManiFlow - a practical framework for sampling from lower dimensional manifolds using \acp{nf}. To this end, we recognised that regions of maximum likelihood in \acp{nf} trained with sufficiently small noise implicitly represent the underlying manifold. Thus, given a trained \ac{nf} we can generate samples on manifold. We proposed two strategies to achieve this.

Firstly, we introduced a general framework based on maximizing the log-likelihood using \eref{eq:optimization_criterion} (LLM). ManiFlow based on LLM is applicable to low-dimensional as well as high-dimensional real world data distributions. It consistently improves 3D point cloud reconstruction (\secref{subsection:pc_autoencoding}) and image generation (\secref{subsection:image_modeling}). In particular, on 3D point cloud autoencoding we find that post-processing samples from \acp{nf} with ManiFlow LLM performs considerably better than AtlasNet while being able to evaluate the likelihood. This was previously not possible~\cite{klokov2020discrete,postels2021go}.

Moreover, we proposed LL-Poisson as a specialized version for 3D point clouds (\secref{subsection:sampling_manifold_poisson}). LL-Poisson utilizes the \ac{nf}'s log-likelihood and its normalized gradients to perform Poisson surface reconstruction. LLM post-processes samples individually. In contrast, LL-Poisson relocates points also based on local information about adjacent points and information about surface normals extracted from the gradient field. Thus, LL-Poisson outperforms LLM on point cloud autoencoding (\secref{subsection:pc_autoencoding}). We showed that the log-likelihood of the \ac{nf} and its gradient improve Poisson surface reconstruction (see \tabref{table:dpm_pp}, \tabref{table:maniflow_poisson_ablation} and \tabref{table:maniflow_poisson_sparse_pointclouds}). We particularly found that the benefit the \ac{nf}'s log-likelihood for LL-Poisson magnifies when operating on sparse point clouds. We conclude that LL-Poisson can also be used to improve mesh creation from sparse point clouds. 

In the realm of 3D shapes, it is also interesting to interpret ManiFlow in the light of implicit neural representations (INRs). INRs represent shapes as the level set of an implicit function parameterized by a \ac{nn} and are typically learned as signed distance functions~\cite{Park_2019_CVPR}. Similarly, ManiFlow naturally encodes the manifold implicitly in its regions of maximum likelihood. Practically the main difference remains that the underlying \ac{nf} in ManiFlow allows direct sampling in the vicinity of the surface whereas an INR requires many evaluations far away from the surface.

While ManiFlow already yields strong improvements over standard \acp{nf}, it simultaneously gives rise to interesting future research avenues. For example, since we have seen that the log-likelihood of \acp{nf} contains geometric information, it is natural to ask whether we can adjust the training of \acp{nf} to improve this property. A resulting approach could lead to the intersection of score-matching and direct likelihood maximization. Furthermore, score matching has recently demonstrated strong performance on point cloud denoising~\cite{luo2021score}. Similarly, it worth investigating whether \acp{nf} equipped with ManiFlow are useful for this task. In fact, such an approach could extend traditional denoising approaches based on kernel density estimates~\cite{schall2005robust}.